\documentclass[conference]{IEEEtran}
\usepackage{graphicx}
\usepackage{dirtytalk}
\graphicspath{ {images/} }

\usepackage{amsmath}

\ifCLASSINFOpdf
\else
\fi
\usepackage{amsmath}
\begin{document}
%
\title{San Francisco Crime Classification}

\author{
\IEEEauthorblockN{
\vspace{.5em}\\
Yehya Abouelnaga}
\IEEEauthorblockA{School of Sciences and Egnineering, Department of Computer Science and Engineering\\
The American University in Cairo, New Cairo 11835, Egypt\\
devyhia@aucegypt.edu\\
\vspace{.25em}\\
}
}


%


\maketitle

\begin{abstract}
San Francisco Crime Classification is an online competition administered by Kaggle Inc. The competition aims at predicting the future crimes based on a given set of geographical and time-based features. In this paper, I achieved a an accuracy that ranks at top \%18, as of May 19th, 2016. I will explore the data, and explain in details the tools I used to achieve that result.
\end{abstract}


%
\IEEEpeerreviewmaketitle

\section{Introduction}
San Francisco first boomed in 1849 during the California Gold Rush, and in the next few decades, the city expanded rapidly both in terms of land area and population. The rapid population increase led to social problems and high crime rate fueled in part by the presence of red-light districts. However, the San Francisco of today is a far cry from its origins as a mining town. San Francisco has seen an influx of technology companies and their workers. While this has resulted in the city being acclaimed as a technological capital, the gentrification of its neighbourhoods have not been entirely well-accepted. It comes as no surprise that a tech-savvy city like San Francisco have decided to publicly release their crime data on their open data platform, and this data is part of an open competition on Kaggle to predict criminal occurrences in the city \cite{Ang2015}.

\section{Dataset Analysis}
The San Francisco Crime Classification dataset contains the following set of features:
\begin{itemize}
\item \textbf{Longitude} -- X coordinates on the map where the crime occurred.
\item \textbf{Latitude} -- Y coordinates on the map where the crime occurred.
\item \textbf{Address} -- The address of the crime incident.
\item \textbf{Day of Week} -- The day of the week (i.e. Thursday)
\item \textbf{Date} -- The date of the crime in the following format: \textit{YYYY-mm-dd hh:MM:ss}. Thus, you can deduce the following: Year, Month, Day, Hour, Minute, Second.
\item \textbf{District} -- Police district to which the crime is assigned.
\item \textbf{Resolution} -- The resolution taken to address the crime.
\item \textbf{Category} -- The type of the crime. This is the target/label that we need to predict.
\end{itemize}
The previous features are provided in the training set. However, in the data set you don't have \textbf{Category}, and \textbf{Resolution} columns. You do have an extra \textbf{Id} column for the purpose of Kaggle submissions.
\par
In our training, we will remove the \textbf{Resolution} column because it's associated with the Category (our label/target). Thus, it's not provided in the test dataset, as well. Now, we will explore the features one by one.

\subsection{Data Distribution}
\subsubsection{Hour}
After experimenting, this turned out to be the most important feature. It provides the highest correlation with the crime category (our target label). This is extracted from the feature \textit{Dates}.
\begin{figure}[!ht]
\centering
\includegraphics[width=0.5\textwidth]{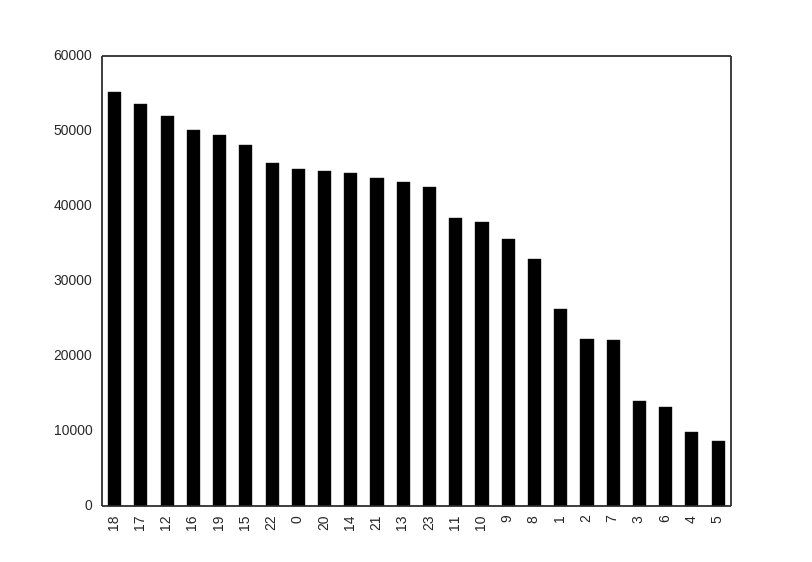}
\caption{Crime Distribution Per Hour}
\end{figure}

\subsubsection{District} There're 10 districts in San Francisco. Some of them are more endowed with crimes than others.
\begin{center}
 \begin{tabular}{||c | c||}
 \hline
 District & Number of Crimes \\ [0.5ex]
 \hline\hline
SOUTHERN & 157,182 \\
\hline
MISSION      & 119,908 \\
\hline
NORTHERN   &   105,296 \\
\hline
BAYVIEW       & 89,431 \\
\hline
CENTRAL      &  85,460 \\
\hline
TENDERLOIN  &   81,809 \\
\hline
INGLESIDE     & 78,845 \\
\hline
TARAVAL      &  65,596 \\
\hline
PARK         &  49,313 \\
\hline
RICHMOND     &  45,209 \\
 \hline
\end{tabular}
\end{center}

\subsubsection{Day Of Week}
Crimes seem to be almost evenly distributed across days of the week. They increase on Fridays, though. Friday night parting culture might have an impact on that spike.
\begin{figure}[!ht]
\centering
\includegraphics[width=0.5\textwidth]{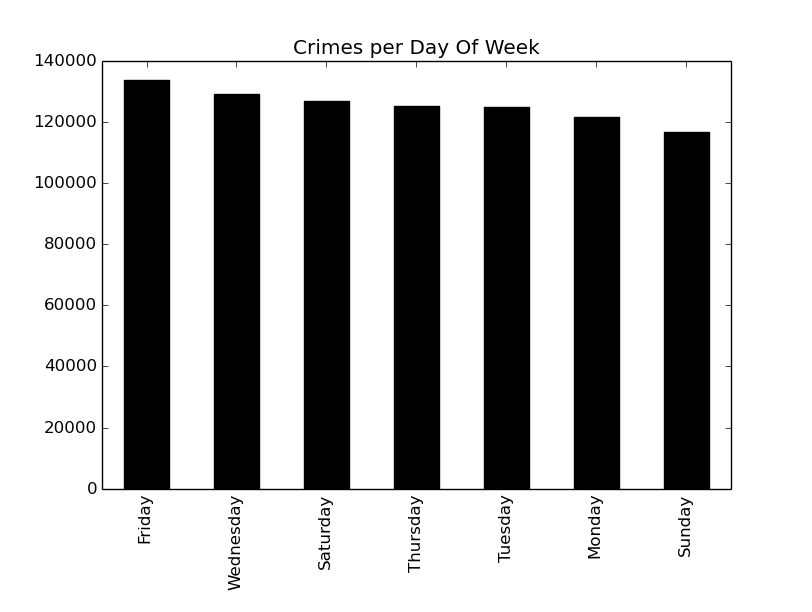}
\caption{Example of a parametric plot ($\sin (x), \cos(x), x$)}
\end{figure}
\subsubsection{Address}
This feature is very sparse among the entire dataset. There are 23,228 unique address in the training dataset. Thus, it's hard to encode these values. The most frequent address is \textit{800 Block of BRYANT ST}. We notice that most of the popular crime places contain the word \say{BLOCK} in them. Thus, when used in classification, it gave better results.
\subsubsection{Category}
This is the target label/crime we want to predict. We've 39 crime categories (i.e. classification classes). The distribution of training sample is very skewed as you can see in the table below.

\begin{center}
 \begin{tabular}{||c | c||}
 \hline
 District & Number of Crimes \\ [0.5ex]
 \hline \hline
LARCENY/THEFT       &           174,900\\
 \hline
OTHER OFFENSES     &           126,182\\
 \hline
NON-CRIMINAL            &        92,304\\
 \hline
ASSAULT                      &   76,876\\
 \hline
DRUG/NARCOTIC        &           53,971\\
 \hline
VEHICLE THEFT            &       53,781\\
 \hline
VANDALISM                   &    44,725\\
 \hline
 ...                  &      ...\\
 \hline
EMBEZZLEMENT           &          1,166\\
 \hline
SUICIDE                        &   508\\
 \hline
FAMILY OFFENSES       &            491\\
 \hline
BAD CHECKS                 &       406\\
 \hline
BRIBERY                        &   289\\
 \hline
EXTORTION                   &      256\\
 \hline
SEX OFFENSES NON FORCIBLE    &     148\\
 \hline
GAMBLING                   &       146\\
 \hline
PORNOGRAPHY/OBSCENE MAT  &          22\\
 \hline
TREA                              &  6\\
 \hline
\end{tabular}
\end{center}
\subsubsection{Street Number}
This is a feature extracted from the Address. About 300,000 of the address have no street number, thus, got a zero value. The remaining 550,000 crimes have street numbers distributed over 80 different numbers. Note that, we had about 26,000 unique addresses. Meanwhile, we have only 80 unique street numbers. This feature increased the classification accuracy.
\subsubsection{Block}
This is another feature extracted from the Address. It only has binary values: block or non-block. There're 617,231 block-based crimes. And, other 260,818 non-block-based crimes. This feature increased accuracy as well.

\section{Evaluation Metric}
The metric used to evaluate the quality of the classifier is the multi-class logarithmic loss, as indicated below.
\[logloss=-\frac{1}{2}\sum_{i=1}^{N}\sum_{j=1}^{M}y_{ij}log(p_{ij})\]

\section{Principal Component Analysis}
In order to increase the classification accuracy, and avoid overfitting, we used PCA to reduce the dimensionality. We noticed that the PCA performs best with the 3 components.
\par
In the following table, we tested the classification accuracy after using PCA, with 1, 2, and 3 components.
\begin{center}
\begin{tabular}{||c | c||}
\hline
\# of Components & Log-loss \\
\hline
1& 2.5103915\\
\hline
2& 2.5094789\\
\hline
3& 2.5056236\\
\hline
\end{tabular}
\end{center}

\section{Classification}
In this section, I will explore different classification models, and compare their accuracy.
\subsection{Features}
We used the Hour, Month, District, Day of Week, Longitude, Latitude, StreetNo, Block, and 3 components of the PCA that preserve the highest variance.

\subsection{$k$-Nearest Neighbors Classifier}
This classifier achieved high results only under high values of $k$.
\begin{center}
\begin{tabular}{||c | c||}
\hline
$k$ & Log-loss (Validation) \\
\hline
39 & 6.341270679\\
100 & 3.966297590\\
500 & 2.837714675\\
1000 & 2.703951916\\
2000 & 2.645995828\\
3000 & 2.628225854\\
4000 & 2.621394788\\
\hline
\end{tabular}
\end{center}

\subsection{XGBClassifier}
This classifier is a very robust regressor. It's very fast, and achieves good results. Here are the results with changes in the \textit{max\_depth}.

\begin{center}
\begin{tabular}{||c | c | c||}
\hline
\textit{max\_depth} & Log-loss (Test) & Log-loss (Validation) \\
\hline
3 & 2.57896 &	2.573599576\\
\hline
4 & 2.57896 &	2.573666882\\
\hline
5 & 2.57428 &	2.568169744\\
\hline
6 & 2.57428 &	2.565219931\\
\hline
\end{tabular}
\end{center}
The results show that the change of depth didn't achieve much better results.

\subsection{Decision Tree Classifier}
Decision Tree Classifiers are very fast compared to the previous ones, and surprisingly, more accurate for this classification problem.
\begin{center}
\begin{tabular}{||c | c||}
\hline
\textit{max\_depth} & Log-loss (Validation) \\
\hline
4 & 2.60898971\\
\hline
5 & 2.592237171\\
\hline
6 & 2.585709647\\
\hline
7 & 2.584480513\\
\hline
8 & 2.597382051\\
\hline
9 & 2.634747421\\
\hline
10 & 2.536769\\
\hline
\end{tabular}
\end{center}

It's noteworthy that \textit{DecisionTreeClassifier} performs best at $max\_depth=10$. We proceeded to experiment with the number of elements in the tree, while fixed $max\_depth=10$. And, we found that the it performs best when $max\_depth=10$, and $n\_elements=256$. This set of parameters yields a log-loss of $2.50849507$.

\subsection{Bayesian Classifier}
This classifier didn't prove to be the best for this set of features. It sets a baseline of $2.64923294$ on the validation set.

\subsection{Random Forrest Classifier}
Random Forrest Classifier proved to be the best for the job. After parameter tuning, at $max\_depth=13$, it performs best. Now, we have to tune the $n\_elements$ parameter.
\begin{center}
\begin{tabular}{||c | c||}
\hline
\textit{n\_elements} & Log-loss (Validation) \\
\hline
10 & 2.441049707\\
\hline
20 & 2.422189106\\
\hline
30 & 2.424566886\\
\hline
40 & 2.420590797\\
\hline
50 & 2.416913452\\
\hline
100 & 2.412998032\\
\hline
150 & 2.411652466\\
\hline
190 & 2.411337273\\
\hline
200 & 2.410574498\\
\hline
200 + StreetNo & \textbf{2.370072457}\\
\hline
200 + StreetNo + Block & \textbf{2.366175731}\\
\hline
\end{tabular}
\end{center}

\section{Kaggle Scores}
After submitting the results to the Kaggle competition, our best classifier placed at 388 out of 2077 (as of May 19, 2016), with a log-loss of \textbf{2.39031}. This is among the \textbf{Top 18\%} on the leaderboard. This result is higher than those achieved in \cite{Ang2015}, \cite{Ke2015}, \cite{Chandrasekar}, and \cite{Haskell2015}.

\section{Future Work}
We still think that we can achieve much higher accuracy when we employ more feature engineering on the Address field. \cite{Papadopoulos2015} suggested that we use Learning by Counting, explained in \cite{Microsoft2016}, in order to generate a log odds feature that might be useful as \cite{Papadopoulos2015} claims. \cite{Papadopoulos2015} also suggested that we use a Neural Network to train on the data. He achieved a much higher accuracy with the use of a Neural Net with 512 hidden layers. We need to experiment with the concept of classifier fusions, mentioned in \cite{Ruta2000}.

\section{Conclusion}
In this paper, I explored a wide spectrum of possible classifiers that might be a good fit for solving the San Francisco Crime Classification problem. We achieved a log-loss metric that was higher than most of the published solutions, with subtle feature engineering, and classifier parameter tuning.


%
%



%

\bibliographystyle{IEEEtran}
\bibliography{Refs}

\begin{thebibliography}{1}
\providecommand{\url}[1]{#1}
\csname url@samestyle\endcsname
\providecommand{\newblock}{\relax}
\providecommand{\bibinfo}[2]{#2}
\providecommand{\BIBentrySTDinterwordspacing}{\spaceskip=0pt\relax}
\providecommand{\BIBentryALTinterwordstretchfactor}{4}
\providecommand{\BIBentryALTinterwordspacing}{\spaceskip=\fontdimen2\font plus
\BIBentryALTinterwordstretchfactor\fontdimen3\font minus
  \fontdimen4\font\relax}
\providecommand{\BIBforeignlanguage}[2]{{%
\expandafter\ifx\csname l@#1\endcsname\relax
\typeout{** WARNING: IEEEtran.bst: No hyphenation pattern has been}%
\typeout{** loaded for the language `#1'. Using the pattern for}%
\typeout{** the default language instead.}%
\else
\language=\csname l@#1\endcsname
\fi
#2}}
\providecommand{\BIBdecl}{\relax}
\BIBdecl

\bibitem{Ang2015}
S.~T. Ang, W.~Wang, and S.~Chyou, ``{San Francisco Crime Classification},''
  \emph{University of California San Diego}, 2015.

\bibitem{Ke2015}
J.~Ke, X.~Li, and J.~Chen, ``{San Francisco Crime Classification},''
  \emph{University of California San Diego}, 2015.

\bibitem{Chandrasekar}
A.~Chandrasekar, A.~S. Raj, and P.~Kumar, ``{Crime Prediction and
  Classification in San Francisco City}.''

\bibitem{Haskell2015}
\BIBentryALTinterwordspacing
C.~Haskell, ``{Kaggle Competition: San Francisco Crime Classification},'' 2015.
  [Online]. Available:
  \url{https://brittlab.uwaterloo.ca/2015/11/01/KaggleSFcrime/}
\BIBentrySTDinterwordspacing

\bibitem{Papadopoulos2015}
\BIBentryALTinterwordspacing
C.~Papadopoulos, ``{Predicting Crime Categories with Address Featurization and
  Neural Nets},'' 2015. [Online]. Available:
  \url{https://www.kaggle.com/c/sf-crime/forums/t/15836/predicting-crime-categories-with-address-featurization-and-neural-nets/103225}
\BIBentrySTDinterwordspacing

\bibitem{Microsoft2016}
\BIBentryALTinterwordspacing
Microsoft, ``{Data Transformation: Learning with Counts},'' 2016. [Online].
  Available: \url{https://msdn.microsoft.com/en-us/library/azure/dn913056.aspx}
\BIBentrySTDinterwordspacing

\bibitem{Ruta2000}
D.~Ruta and B.~Gabrys, ``{An Overview of Classifier Fusion Methods},''
  \emph{Computing and Information Systems}, vol.~7, pp. 1--10, 2000.

\end{thebibliography}

\end{document}